\pdfoutput=1

\documentclass[11pt]{article}

\usepackage{ACL2023} 

\usepackage{times}
\usepackage{latexsym}

\usepackage[T1]{fontenc}

\usepackage[utf8]{inputenc}

\usepackage{microtype}

\usepackage{inconsolata}
\usepackage{graphicx}
\usepackage{amsmath}
\usepackage{listings}
\usepackage{tabularx}
\usepackage{subcaption}
\usepackage{float}

\title{PetKaz at SemEval-2024 Task 3: \\ Advancing Emotion Classification with an LLM for \\ Emotion-Cause Pair Extraction in Conversations}

\author{Roman Kazakov, Kseniia Petukhova, Ekaterina Kochmar\\
  Mohamed bin Zayed University of Artificial Intelligence \\
  \texttt{\{roman.kazakov, kseniia.petukhova, ekaterina.kochmar\}@mbzuai.ac.ae} \\}

\begin{document}
\maketitle
\begin{abstract}
In this paper, we present our submission to the SemEval-2023 Task~3 ``The Competition of Multimodal Emotion Cause Analysis in Conversations'',  focusing on extracting emotion-cause pairs from dialogs. Specifically, our approach relies on combining fine-tuned {\tt GPT-3.5} for emotion classification and a BiLSTM-based neural network to detect causes. We score 2nd in the ranking for Subtask 1, demonstrating the effectiveness of our approach through one of the highest weighted-average proportional $F_1$ scores recorded at 0.264. Our code is available at \url{https://github.com/sachertort/petkaz-semeval-ecac}.
\end{abstract}

\section{Introduction}
\label{s:intro}
\begin{figure*}
  \centering
  \includegraphics[width=\linewidth]{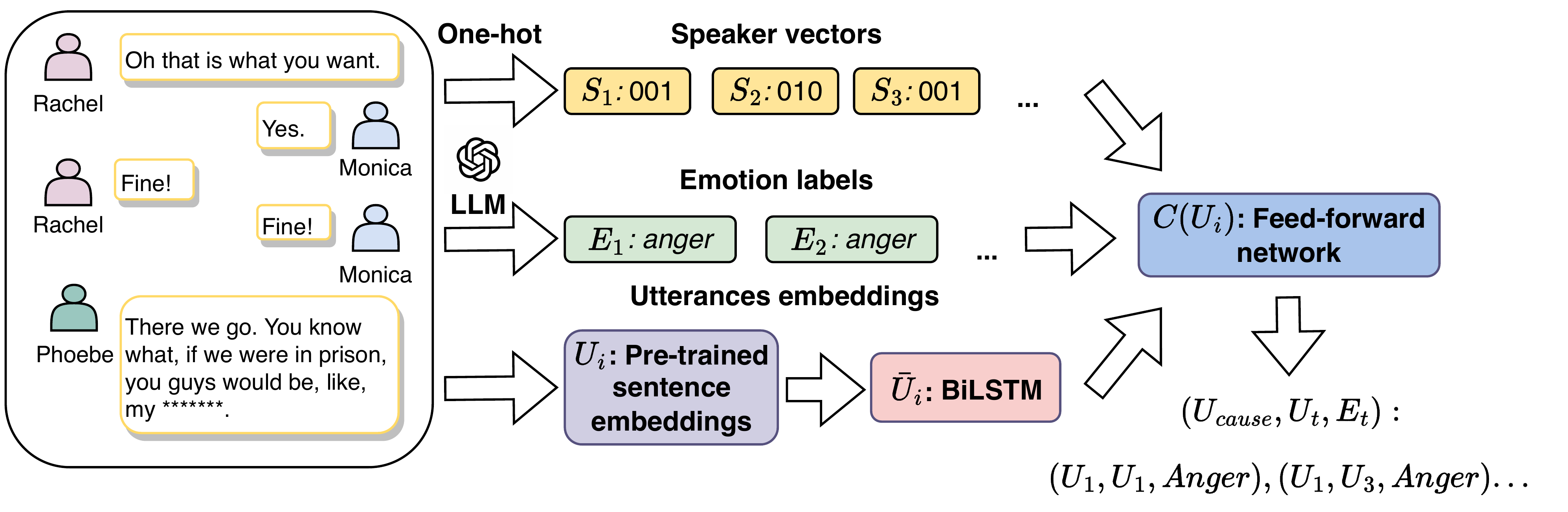}
  \caption{The pipeline for ECPE. Utterances are classified with emotion labels $E_i$, and speakers are represented with one-hot vectors $S_i$. Utterances are then encoded with pre-trained sentence embeddings $U_i$ and enriched with context by BiLSTM $\bar{U}_i$. For each target utterance $U_t$, we detect whether any other utterance from the conversation history $H(U_t)$ is causal using a feed-forward network. $\bar{U}_i$, $S_i$ (of a potential causal utterance), $\bar{U}_t$, $S_t$, and $E_t$ are concatenated, and then, binary classification is performed. The pipeline outputs labelled emotion-cause pairs $(U_i, U_t, E_t)$.}
  \label{fig:scheme}
\end{figure*}

Developing dialog systems is a complex task that has attracted considerable attention from many technology companies and universities over the last 70 years since the introduction of Eliza in 1966~\citep{weizenbaum1966eliza}. Modern large language models (LLMs) like GPT-4~\citep{openai2023gpt4} are trained to avoid causing harm and often assert their lack of personal opinions on intricate matters, which is not at all natural for conversations. They do not respond in a way that is truly human, and they do not understand the range of feelings that words can cause. Recognizing the emotional implications of an utterance provides a deeper understanding of dialog, enabling the development of more human-like dialog systems. These systems could navigate conversations using a comprehensive understanding of emotional dynamics and planning responses based on this understanding rather than just predicting likely outcomes.

To bridge the gap between machine-generated dialogs and rich, complex human communication, we develop models for SemEval-2024 Task 3 ``The Competition of Multimodal Emotion Cause Analysis in Conversations''\footnote{\url{https://nustm.github.io/SemEval-2024\_ECAC/}} (ECAC)~\cite{wang-EtAl:2024:SemEval20244}. This task was previously introduced in~\citet{xia2019emotion} and later in~\citet{wang2023multimodal}, where the authors also described a multimodal dataset called {\em Emotion-Cause-in-Friends} (ECF) for this task.

We focus only on Subtask 1, ``Textual Emotion-Cause Pair Extraction in Conversations'' (ECPE),\footnote{We did not participate in the multimodal track.} where the goal is to classify emotions and extract the corresponding textual causal spans. To accomplish this, we propose a two-stage pipeline: (1) first, emotions are classified using a fine-tuned LLM, and then (2) causes are extracted with a simple neural network consisting of BiLSTM and linear layers (see Figure~\ref{fig:scheme}). Our system achieved a weighted-average proportional $F_1$ score of 0.264, the primary metric in this competition's evaluation phase on the test set. Consequently, our team ranked 2nd out of 15 participating teams based on this metric. We provide an extensive analysis of the model's performance in Section~\ref{s:analysis}.

\section{Related Work}
\label{s:related-work}




Recent research in the field of dialog systems and emotion-cause extraction has seen significant advancements through various innovative approaches, some of which we overview in this section. For instance, \citet{chen-etal-2023-enhance} introduce a novel technique that uses graphs to model ``causal skeletons'' alongside a causal autoencoder (CAE) for refining these models by integrating both implicit and explicit causes.

Following closely, \citet{zhang-etal-2023-dualgats} present Dual Graph Attention Networks (DualGATs) that leverage discourse structure and speaker context through a combination of Discourse-aware GAT (DisGAT) and Speaker-aware GAT (SpkGAT), enriched with an interaction module for effective information exchange and context capturing.

Moving to earlier work, \citet{kong2022discourse} propose a discourse-aware model (DAM) that integrates emotion cause extraction with discourse parsing, using a Gated Graph Neural Network (GNN) to encode discourse structures and conversation features within a multi-task learning framework, enhancing the understanding of conversational context and structure.

Finally, \citet{gao-etal-2021-improving-empathetic} focus on improving dialog systems' empathetic response generation by identifying emotion causes. Their framework combines an emotion reasoner for predicting emotion and its cause with a response generator that employs a gated attention mechanism to emphasize important words, exploring both hard and soft gating strategies.

\section{System Overview}
\label{s:methodology}

Our pipeline consists of two stages. Specifically, to identify emotion-cause pairs and emotion types, dialogs are  passed through the following modules:
\begin{enumerate}
    \item classification of utterances with emotion types (including \textit{neutral} for non-emotional utterances) with a supervised fine-tuned LLM; and \vspace{-0.5em}
    \item extraction of cause utterances with a BiLSTM-based network.
\end{enumerate}

The full pipeline is shown in Figure~\ref{fig:scheme}. Due to the limitations of the data, we perform the tasks separately, and we elaborate on each of the stages in the following sections.

\subsection{Emotion classification}
To categorize an utterance with an emotion label $E_t$, within our pipeline an LLM should consider both the target utterance $U_t$, which is the $t$\textsuperscript{th} utterance in a conversation, and the preceding utterance $U_{t-1}$. It is especially important when we deal with very short turns, such as ``Instead of... ?'', ``No.'', ``Yeah, maybe...''. Indeed, it would be more accurate to utilize causal utterances rather than antecedent ones; however, at the initial stage, these are unknown to us, necessitating the use of a meaningful alternative.

\begin{figure}
  \centering
  \includegraphics[width=\linewidth]{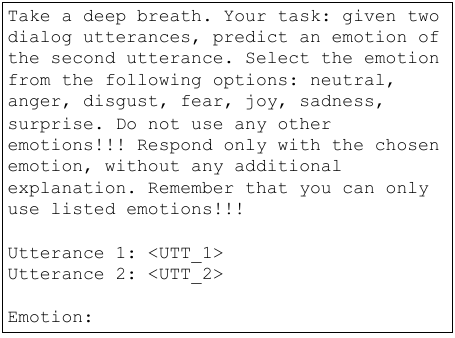}
  \caption{The prompt used to perform emotion classification with {\tt GPT-3.5}.}
  \label{fig:prompt}
\end{figure}

For this stage, we fine-tune \texttt{GPT-3.5}.\footnote{\texttt{gpt-3.5-turbo-1106}: \url{https://platform.openai.com/docs/models/gpt-3-5-turbo}} As a \texttt{system}'s input, we provide the prompt consisting of an instruction, $U_{t-1}$ (\texttt{<UTT\_1>}), and $U_t$ (\texttt{<UTT\_2>}) as is shown in Figure~\ref{fig:prompt}.
This particular prompt was selected during the preliminary prompt engineering stage. The \texttt{assistant}'s output consists of one word -- the emotion type. 

We also note that preliminary experiments showed that the LLM performed poorly in zero-shot and few-shot settings on the emotion detection task, at least on the ECF dataset (see Section~\ref{s:emclass} and Table~\ref{tab:metrics}). Therefore, we had to fine-tune it.

\subsection{Cause extraction}

The second stage is concerned with the detection of the causal utterances for non-emotional utterances in a binary way. Let the whole conversational history of an utterance $U_t$ be $H(U_t) = [U_1, U_t]$, then the set of all causal utterances is $C(U_t) \subseteq H(U_t)$. In addition, speakers are encoded with one-hot vectors $S_1... S_n$ within each dialog.

First, we need to enrich utterance embeddings $U_1... U_n$\footnote{We use the same notation for utterances and their embeddings for simplification purposes.} obtained from a pre-trained model with the context within the conversation. Bidirectional LSTM~\cite{hochreiter1997long} was chosen because it can preserve context information in sequential settings using the content of the previous hidden state in encoding the current one. This way, we get new utterance representations  $\bar{U}_1... \bar{U}_n$.

Then, for each target utterance $U_t$ with $E_t~\neq~neutral$, we construct $t$ representations:
\begin{equation}
    \bar{U}_{i} \mathbin\Vert S_{i} \mathbin\Vert \bar{U}_{t} \mathbin\Vert S_{t} \mathbin\Vert E_t, \forall i \in [1, ..., t]
\end{equation} 
containing one of the previous utterances or the target utterance embedding itself $\bar{U}_{i}$ as a potential cause, its speaker vector $S_{i}$, the target utterance embedding $\bar{U}_{t}$, its speaker vector $S_{t}$, and the emotion label $E_t$. We pass them to a feed-forward neural network and obtain binary predictions $\{0, 1\}$, where $1$ means that $U_{i}$ is a causal utterance and 0 stands for the opposite. All $U_{i}$ for which 1 is predicted make up $C(U_t)$.
Thus, for each $U_t$ with $E_t$ $\neq$ $neutral$ we obtain from 0 to $t$ labelled emotion-cause pairs $(U_i, U_t, E_t), \text{where } U_i \in C(U_t)$, consisting of the causal utterance,\footnote{We did not extract causal spans and used the whole causal utterance in the evaluation.} the emotion utterance, and the emotion label.

We have decided not to extract specific spans from the utterances classified as causes, following a thorough review of the dataset. This decision is based on our observation that these spans often defy straightforward explanations, even from a human annotator perspective. 
Here are some examples, where the rationale behind the spans remains unclear to us: 
\begin{itemize}
    \item The final punctuation marks are often not included in the cause span: e.g., while the complete utterance is \textit{Instead of [...]?}, the identified cause span is \textit{Instead of [...]}
    \item For the statement \textit{Me, I ... I went for the watch}, the span is \textit{I went for the watch} 
    \item For the sentence \textit{You know you probably did not know this, but back in high school, I had a, um, major crush on you}, the cause span is defined as \textit{you probably did not know this, but back in high school, I had a, um, major crush on you}
\end{itemize}

We believe that this part of the task can be more accurately defined as a causal emotion entailment~\citep{poria2021recognizing}. Additionally, we note that there is an inconsistency in the dataset's annotation: specifically, the task organizers define emotion causes by identifying specific spans within an utterance, yet the emotional responses themselves are treated as consisting of entire utterances. 
For these reasons, we have decided that it would be methodologically more appropriate to omit the exact span detection step from our pipeline.

\section{Experimental Setup}
\label{s:experimental-setup}

\subsection{Data}

The dataset proposed for the shared task contains conversations from the \textit{Friends} series annotated with emotion-cause pairs and emotion labels, including \textit{anger}, \textit{disgust}, \textit{fear}, \textit{joy}, \textit{sadness}, \textit{surprise} from~\citet{ekhman1987}, or \textit{neutral} for non-emotional utterances.

The shared task organizers highlight that $91\%$ of emotions have corresponding causes and one emotion may be triggered by multiple causes in different utterances. In addition, we have noticed that $16\%$ of them cause several different emotions.

The organizers did not provide a standalone development set, so we had to split the training set ourselves using a ratio of 9:1 relative to the dialogs. The final data splits are shown in Table~\ref{tab:split}. 

\begin{table}[t]
    \tiny
    \centering
    \resizebox{\columnwidth}{!}{
    \begin{tabular}{lrrr} \hline 
         \textbf{Set} & \textbf{\# dialogs}& \textbf{\# utterances} & \textbf{\# EC} \\ \hline 
         Training & 1,236 & 12,346 & 8,565 \\ 
         Development & 138 & 1,273 & 799 \\ \hline
         Total & 1,374 & 13,619 & 9,364 \\ \hline
    \end{tabular}}
    \caption{Distribution of dialogs, utterances, and emotion-cause pairs (``EC'') across the split sets.}
    \label{tab:split}
\end{table}

\begin{table*}[t]
    \tiny
    \centering
    \resizebox{\linewidth}{!}{
    \begin{tabular}{lrrrrrrrrr} \hline
         \textbf{Approach} & \textbf{\textit{neutral}} & \textbf{\textit{anger}} & \textbf{\textit{disgust}} & \textbf{\textit{fear}} & \textbf{\textit{joy}} & \textbf{\textit{sadness}} & \textbf{\textit{surprise}} & \textbf{macro} & \textbf{w-avg.}\\ \hline
         Zero-shot & 0.61 & 0.43 & 0.30 & 0.32 & 0.54 & 0.47 & 0.50 & 0.45 & 0.54 \\
         Few-shot & 0.57 & 0.49 & 0.31 & 0.34 & 0.54 & 0.37 & 0.41 & 0.43 & 0.51 \\
         Fine-tuning & \textbf{0.70} & \textbf{0.57} & \textbf{0.42} & \textbf{0.51} & \textbf{0.63} & \textbf{0.52} & \textbf{0.66} & \textbf{0.57} & \textbf{0.64} \\ \hline
    \end{tabular}
    }
    \caption{$F_1$ scores on emotion classification with GPT-3.5 across different approaches.}
    \label{tab:metrics}
\end{table*}

\subsection{Training and architecture details}
We fine-tune {\tt GPT-3.5} with the default hyperparameters recommended by OpenAI\footnote{\url{https://platform.openai.com/docs/api-reference/fine-tuning/create}} using two epochs, which is the number automatically chosen by the platform.

The cause extractor model is initialized with mean pooling from the penultimate layer's hidden state of the pre-trained  \texttt{bert-base-uncased}.\footnote{\url{https://huggingface.co/bert-base-uncased}} Our neural network consists of three BiLSTM layers, one hidden linear layer accompanied by batch normalization, and a ReLU activation function. 

For training, we employ the Adam optimizer with the learning rate of 1$e$-4, weight decay ($L_2$-norm regularization) of 1$e$-5, and cross-entropy as the loss function. We train the model for 200 epochs using a batch size of 32.

As a framework for training and evaluation, we use \texttt{PyTorch}\footnote{\url{https://pytorch.org}}~\citep{paszke2019pytorch}. 

\subsection{Evaulation measures}

As proposed in the shared task, we apply the weighted average (w-avg.) $F_1$ score by emotion type for evaluation. The specific implementation of $F_1$ score for the ECPE task can be found in~\citet{xia-ding-2019-emotion}. In this setting, an emotion-cause pair is considered as correctly predicted if the index of an emotion utterance, an emotion type, and the index of the cause utterance match the entry in the gold dataset. There are two strategies related to causal span detection: {\em strict} $F_1$ (the same span) and {\em proportional} $F_1$ (overlap).\footnote{For the details on the metrics, refer to \url{https://github.com/NUSTM/SemEval-2024_ECAC/tree/main/CodaLab/evaluation}.}

\section{Results}
\label{s:results-discussion}

Our final submission was evaluated on the test set and achieved the following results: 
\begin{itemize}
    \item w-avg. proportional $F_1$: 0.264; \vspace{-0.5em}
    \item w-avg. strict $F_1$: 0.104.
\end{itemize}
As a result, we score second out of fifteen teams participating in Subtask 1 according to the main shared task metric -- w-avg. proportional $F_1$.

\subsection{Emotion classification performance}
\label{s:emclass}

\begin{figure}[t]
  \centering
  \includegraphics[width=\linewidth]{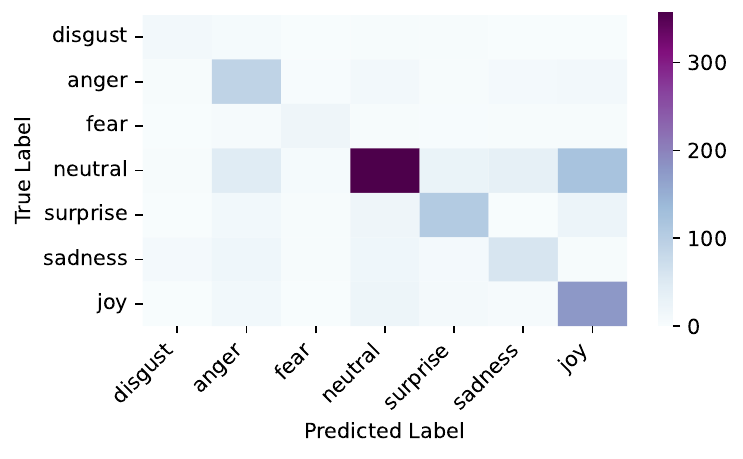}
  \caption{Performance of our emotion classifier on our development set.}
  \label{fig:cm}
\end{figure}

\begin{figure}[t]
  \centering
  \includegraphics[width=\linewidth]{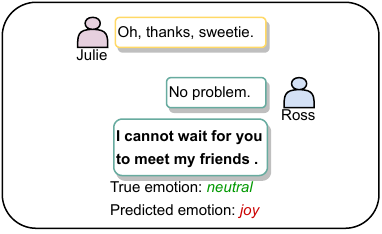}
  \caption{An example of a dialog where our model classified neutral utterance as \textit{joy}.}
  \label{fig:dialog}
\end{figure}

\begin{figure*}[t]
     \centering
     \begin{subfigure}[b]{0.49\textwidth}
         \centering
         \includegraphics[width=\textwidth]{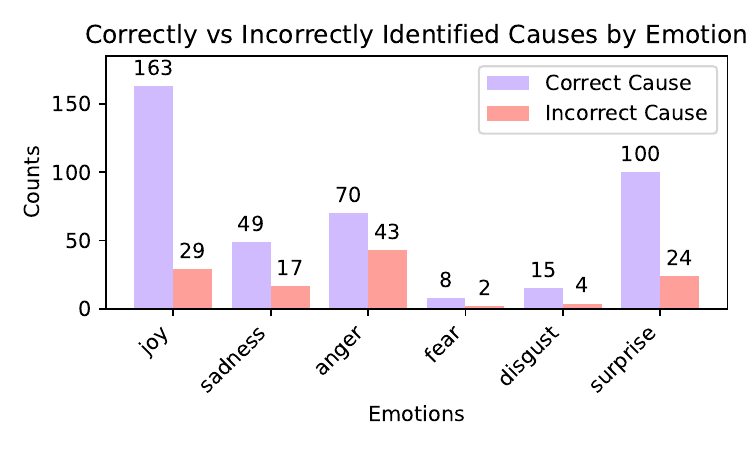}
         \caption{Analysis across emotions on our development set (on correctly identified emotions only).}
         \label{fig:cause_by_emotion}
     \end{subfigure}
     \hfill
     \begin{subfigure}[b]{0.49\textwidth}
         \centering
         \includegraphics[width=\textwidth]{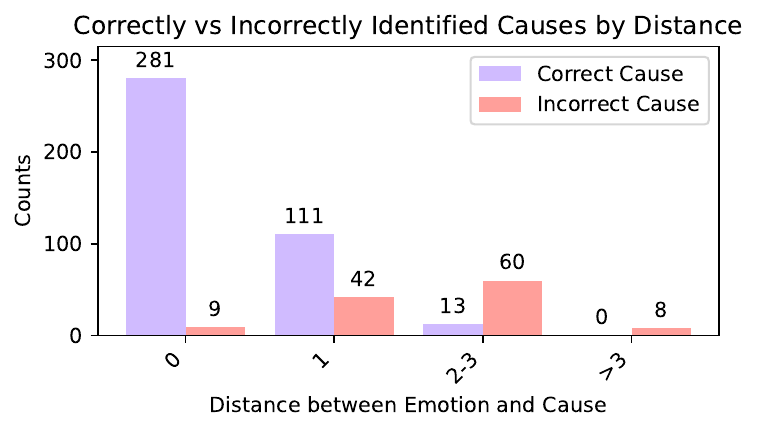}
         \hspace{1 mm}
         \caption{Break-down of distance between emotion and cause on our development set (on correctly identified emotions only).}
         \label{fig:distance}
     \end{subfigure}
    \caption{Performance of the cause extractor.}
    \label{fig:sub}
\end{figure*}

\begin{figure}[t]
  \centering
  \includegraphics[width=\linewidth]{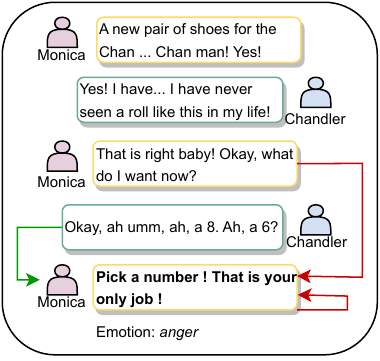}
  \caption{An example of a dialog with annotated causes for \textit{anger} (green for causes correctly identified by our model, and red for causes that our model failed to recognize).}
  \label{fig:anger}
\end{figure}

Table~\ref{tab:metrics} overviews the performance of emotion classification using {\tt GPT-3.5} across different paradigms: zero-shot, few-shot, and fine-tuning. We note that zero- and few-shot settings use the same prompt (see Figure~\ref{fig:prompt}), with the few-shot setting including one handpicked example per each emotion type (see Appendix~\ref{ap:fs}). As expected, fine-tuning yields the best results on all emotion types and overall. Interestingly, few-shot prompting performs worse than zero-shot, which suggests that examples hamper the model's understanding of emotion types instead of improving it. 

Utterances of \textit{disgust} type turn out to be the most difficult to predict correctly: one of the possible reasons is that they are insufficiently represented in the training set (amounting to only about 6\% of emotional utterances). However, the zero-shot and few-shot settings also show the poorest performance on \textit{disgust}.

Our analysis of the emotion classifier's performance across different emotion types shows that the model often incorrectly classifies \textit{neutral} utterances as indicative of \textit{joy} (see Figure~\ref{fig:cm}). After further investigation, we have found that a large number of these incorrectly categorized cases contain greetings (``\textit{Hi!}'') and expressions of gratitude (``\textit{Thank you!}'', ``\textit{You're welcome!}), which, according to our dataset, should be \textit{neutral}, yet our classifier interprets them as \textit{joy}. This implies that text alone may not be enough to identify an emotion, given that such utterances can express joy or remain emotionally neutral. There are other controversial cases, such as a conversation between two lovers shown in Figure~\ref{fig:dialog}, where the statement ``\textit{I cannot wait for you to meet my friends}'' is actually more likely to express joy rather than neutrality.

\subsection{Analysis}
\label{s:analysis}



We also evaluate our model on its ability to identify the causes of utterances expressing different emotions, as shown in Figure~\ref{fig:cause_by_emotion}. Based on this analysis, the greatest challenge for our model is determining causes of \textit{anger}. Similarly, manual analysis shows that this task is difficult for humans as well. As an example, Figure~\ref{fig:anger} highlights a scenario where the source of anger in Monica's utterance is not only attributed to the preceding utterance from Chandler but is also caused by the utterance that came before Chandler's, as well as the context of Monica's own statement. Intricacies like this one highlight the controversies present in the dataset.


Additionally, we have looked into how well our model performs in determining the emotion's cause based on how close it is to the emotional utterance, as we show in Figure~\ref{fig:distance}. First of all, it transpires that most emotional utterances are self-caused. Furthermore, our analysis shows that there is a clear correlation between the cause's distance from the emotional utterance and our model's identification accuracy: the further away the cause, the lower the model's performance.


In the course of our analysis, we have discovered instances where emotions appear before their causes. This observation suggests that the organizers' definition of a cause in dialog contexts is non-trivial, as, typically, we would expect that something happens and triggers an emotion. However, in the case of the preceding emotion, the cause is fundamentally different: it is a reason in terms of linguistics and it explains the emotion, but it does not trigger it (for the difference between \textsc{Cause} and \textsc{Reason}, please refer to~\citealp{Ruppenhofer2006FrameNetIE}).


Overall, accurate identification of emotions and their causes within utterances proves to be a complex challenge, not only for models but also for humans. All issues mentioned above point to important problems in the dataset that need to be carefully thought through and fixed to enhance both the accuracy and reliability of ECPE efforts.

\section{Conclusions}
\label{s:conclusion}
Our work presents a novel approach to emotion-cause pair extraction in conversations, using the capabilities of an LLM (specifically, {\tt GPT-3.5}) for emotion classification. This methodology is further enhanced by the use of a BiLSTM-based neural network for extracting causes. Our system outperforms most of the submissions to the shared task, scoring 2nd in the overall ranking according to the main metric of weighted-average proportional $F_1$. For future enhancements to our pipeline, we consider the following improvements:\vspace{-0.3em}
\begin{itemize}
    \item Firstly, data annotation itself can be expanded and improved, potentially via the use of an LLM for annotation. \vspace{-0.5em}
    \item Secondly, speaker representations can be improved to enhance the understanding and processing of the dialogs. \vspace{-0.5em}
    \item Finally, more accurate methods of LLM-based cause extraction can be developed further.
\end{itemize}

\section*{Limitations}
Due to OpenAI's policy, we are unable to share our fine-tuned model. Therefore, those wishing to reproduce our experiments will need to do the fine-tuning independently. Overall, the usage of an open-source solution instead of a proprietary LLM can be one of the future directions. Also, it may be applied using a more specific framework like InstructERC~\citep{lei2024instructerc}.

Furthermore, our research is limited to the emotions present in the provided task data. Consequently, adding new emotions would require further fine-tuning. Due to the shared task rules, we have to develop our system based only on the presented dataset that is limited to a single concrete domain (\textit{Friends} series) and the English language.

\section*{Acknowledgements}

We are grateful to Mohamed bin Zayed University of Artificial Intelligence (MBZUAI) for supporting this work. We also thank the anonymous reviewers for their valuable feedback.

\bibliography{anthology,custom}

\begin{thebibliography}{16}
\expandafter\ifx\csname natexlab\endcsname\relax\def\natexlab#1{#1}\fi

\bibitem[{Chen et~al.(2023)Chen, Yang, Luo, and Zhu}]{chen-etal-2023-enhance}
Hang Chen, Xinyu Yang, Jing Luo, and Wenjing Zhu. 2023.
\newblock \href {https://doi.org/10.18653/v1/2023.emnlp-main.33} {How to enhance causal discrimination of utterances: A case on affective reasoning}.
\newblock In \emph{Proceedings of the 2023 Conference on Empirical Methods in Natural Language Processing}, pages 494--512, Singapore. Association for Computational Linguistics.

\bibitem[{Ekman et~al.(1987)Ekman, Friesen, O'Sullivan, Chan, Diacoyanni-Tarlatzis, Heider, Krause, LeCompte, Pitcairn, and Ricci~Bitti}]{ekhman1987}
Paul Ekman, Wallace Friesen, Maureen O'Sullivan, A.~Chan, Irene Diacoyanni-Tarlatzis, Karl Heider, Rainer Krause, William LeCompte, Tom Pitcairn, and Pio Ricci~Bitti. 1987.
\newblock \href {https://doi.org/10.1037/0022-3514.53.4.712} {Universals and cultural differences in the judgments of facial expressions of emotion}.
\newblock \emph{Journal of personality and social psychology}, 53:712--7.

\bibitem[{Gao et~al.(2021)Gao, Liu, Deng, Wang, Cao, Du, and Xu}]{gao-etal-2021-improving-empathetic}
Jun Gao, Yuhan Liu, Haolin Deng, Wei Wang, Yu~Cao, Jiachen Du, and Ruifeng Xu. 2021.
\newblock \href {https://doi.org/10.18653/v1/2021.findings-emnlp.70} {Improving empathetic response generation by recognizing emotion cause in conversations}.
\newblock In \emph{Findings of the Association for Computational Linguistics: EMNLP 2021}, pages 807--819, Punta Cana, Dominican Republic. Association for Computational Linguistics.

\bibitem[{Hochreiter and Schmidhuber(1997)}]{hochreiter1997long}
Sepp Hochreiter and Jürgen Schmidhuber. 1997.
\newblock \href {https://doi.org/10.1162/neco.1997.9.8.1735} {Long short-term memory}.
\newblock \emph{Neural computation}, 9:1735--80.

\bibitem[{Kong et~al.(2022)Kong, Yu, Yuan, Fu, and Gong}]{kong2022discourse}
Dexin Kong, Nan Yu, Yun Yuan, Guohong Fu, and Chen Gong. 2022.
\newblock Discourse-aware emotion cause extraction in conversations.
\newblock \emph{arXiv preprint arXiv:2210.14419}.

\bibitem[{Lei et~al.(2024)Lei, Dong, Wang, Wang, and Wang}]{lei2024instructerc}
Shanglin Lei, Guanting Dong, Xiaoping Wang, Keheng Wang, and Sirui Wang. 2024.
\newblock \href {http://arxiv.org/abs/2309.11911} {Instruct{ERC}: Reforming emotion recognition in conversation with a retrieval multi-task {LLM}s framework}.

\bibitem[{OpenAI(2023)}]{openai2023gpt4}
OpenAI. 2023.
\newblock \href {http://arxiv.org/abs/2303.08774} {{GPT}-4 technical report}.

\bibitem[{Paszke et~al.(2019)Paszke, Gross, Massa, Lerer, Bradbury, Chanan, Killeen, Lin, Gimelshein, Antiga, Desmaison, Köpf, Yang, DeVito, Raison, Tejani, Chilamkurthy, Steiner, Fang, Bai, and Chintala}]{paszke2019pytorch}
Adam Paszke, Sam Gross, Francisco Massa, Adam Lerer, James Bradbury, Gregory Chanan, Trevor Killeen, Zeming Lin, Natalia Gimelshein, Luca Antiga, Alban Desmaison, Andreas Köpf, Edward Yang, Zach DeVito, Martin Raison, Alykhan Tejani, Sasank Chilamkurthy, Benoit Steiner, Lu~Fang, Junjie Bai, and Soumith Chintala. 2019.
\newblock \href {http://arxiv.org/abs/1912.01703} {{PyTorch}: An imperative style, high-performance deep learning library}.

\bibitem[{Poria et~al.(2021)Poria, Majumder, Hazarika, Ghosal, Bhardwaj, Jian, Hong, Ghosh, Roy, Chhaya et~al.}]{poria2021recognizing}
Soujanya Poria, Navonil Majumder, Devamanyu Hazarika, Deepanway Ghosal, Rishabh Bhardwaj, Samson Yu~Bai Jian, Pengfei Hong, Romila Ghosh, Abhinaba Roy, Niyati Chhaya, et~al. 2021.
\newblock Recognizing emotion cause in conversations.
\newblock \emph{Cognitive Computation}, 13:1317--1332.

\bibitem[{Ruppenhofer et~al.(2006)Ruppenhofer, Ellsworth, Petruck, Johnson, and Scheffczyk}]{Ruppenhofer2006FrameNetIE}
Josef Ruppenhofer, Michael Ellsworth, Miriam~R.L. Petruck, Christopher~R. Johnson, and Jan Scheffczyk. 2006.
\newblock \emph{{FrameNet II}: Extended Theory and Practice}.
\newblock International Computer Science Institute, Berkeley, California.
\newblock Distributed with the FrameNet data.

\bibitem[{Wang et~al.(2023)Wang, Ding, Xia, Li, and Yu}]{wang2023multimodal}
Fanfan Wang, Zixiang Ding, Rui Xia, Zhaoyu Li, and Jianfei Yu. 2023.
\newblock Multimodal emotion-cause pair extraction in conversations.
\newblock \emph{IEEE Transactions on Affective Computing}, 14(3):1832--1844.

\bibitem[{Wang et~al.(2024)Wang, Ma, Xia, Yu, and Cambria}]{wang-EtAl:2024:SemEval20244}
Fanfan Wang, Heqing Ma, Rui Xia, Jianfei Yu, and Erik Cambria. 2024.
\newblock \href {https://aclanthology.org/2024.semeval2024-1.273} {Semeval-2024 task 3: Multimodal emotion cause analysis in conversations}.
\newblock In \emph{Proceedings of the 18th International Workshop on Semantic Evaluation (SemEval-2024)}, pages 2022--2033, Mexico City, Mexico. Association for Computational Linguistics.

\bibitem[{Weizenbaum(1966)}]{weizenbaum1966eliza}
Joseph Weizenbaum. 1966.
\newblock {ELIZA}—a computer program for the study of natural language communication between man and machine.
\newblock \emph{Communications of the ACM}, 9(1):36--45.

\bibitem[{Xia and Ding(2019{\natexlab{a}})}]{xia2019emotion}
Rui Xia and Zixiang Ding. 2019{\natexlab{a}}.
\newblock Emotion-cause pair extraction: A new task to emotion analysis in texts.
\newblock In \emph{Proceedings of the 57th Annual Meeting of the Association for Computational Linguistics}, pages 1003--1012.

\bibitem[{Xia and Ding(2019{\natexlab{b}})}]{xia-ding-2019-emotion}
Rui Xia and Zixiang Ding. 2019{\natexlab{b}}.
\newblock \href {https://doi.org/10.18653/v1/P19-1096} {Emotion-cause pair extraction: A new task to emotion analysis in texts}.
\newblock In \emph{Proceedings of the 57th Annual Meeting of the Association for Computational Linguistics}, pages 1003--1012, Florence, Italy. Association for Computational Linguistics.

\bibitem[{Zhang et~al.(2023)Zhang, Chen, and Chen}]{zhang-etal-2023-dualgats}
Duzhen Zhang, Feilong Chen, and Xiuyi Chen. 2023.
\newblock \href {https://doi.org/10.18653/v1/2023.acl-long.408} {{D}ual{GAT}s: Dual graph attention networks for emotion recognition in conversations}.
\newblock In \emph{Proceedings of the 61st Annual Meeting of the Association for Computational Linguistics (Volume 1: Long Papers)}, pages 7395--7408, Toronto, Canada. Association for Computational Linguistics.

\end{thebibliography}
\bibliographystyle{acl_natbib}

\appendix

\onecolumn

\section{Prompt for Few-shot}
\label{ap:fs}
\begin{figure*}[!ht]
  \centering
  \includegraphics[width=\linewidth]{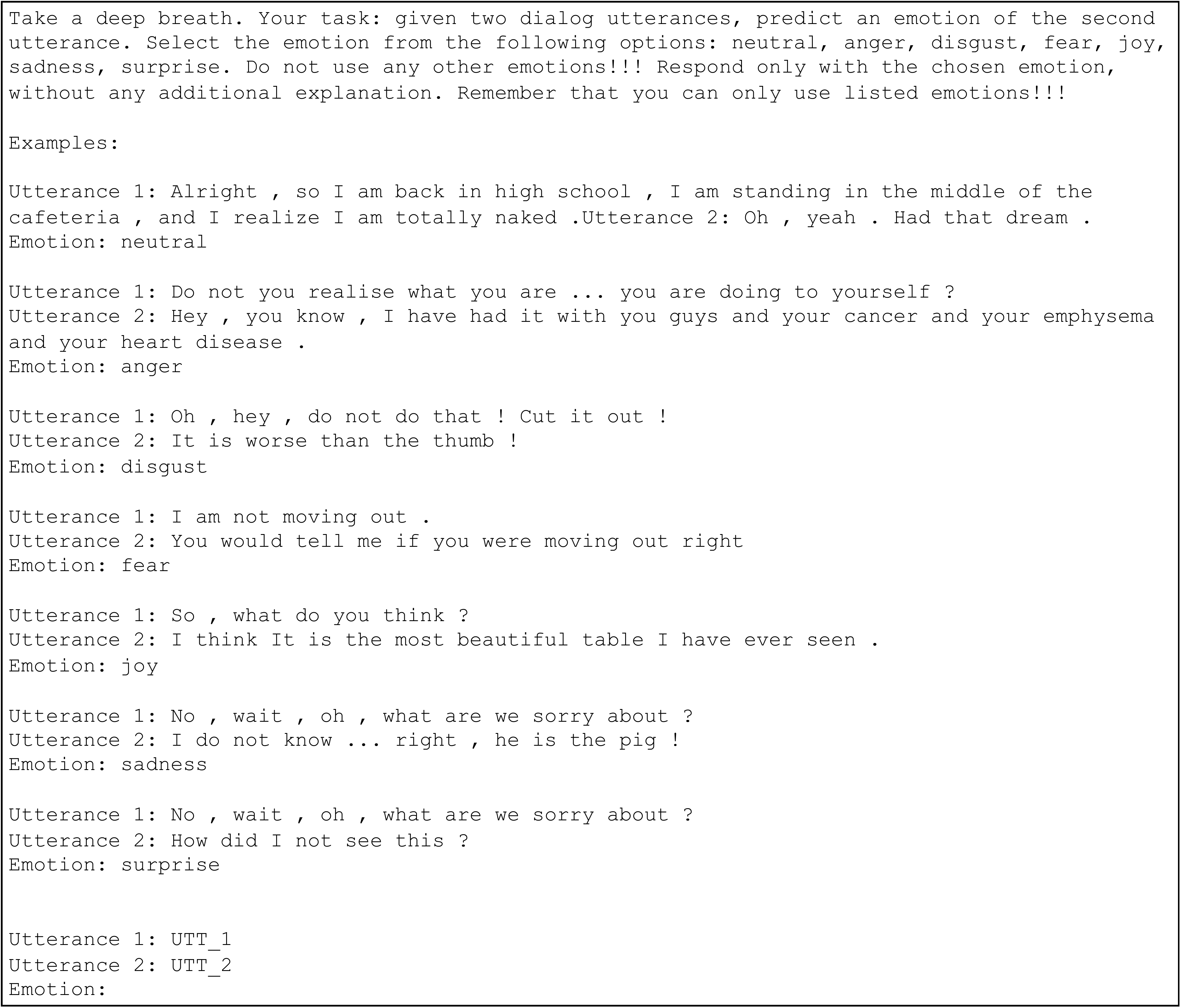}
  \caption{The prompt used to perform emotion classification with {\tt GPT-3.5} in the few-shot setting.}
  \label{fig:prompt_fs}
\end{figure*}

\end{document}